\documentclass[USenglish,twocolumn]{article}
\usepackage[utf8]{inputenc}
\usepackage[big,online]{dgruyter}
\usepackage[sort&compress,square,numbers]{natbib}
\usepackage{times}
\usepackage{float} 
\usepackage[dvipsnames]{xcolor}
\begin{document}

  \DOI{/10.1515/cdbme-2022-0009}
  \openaccess
  \pagenumbering{gobble}

\title{Data-Efficient Vision Transformers for Multi-Label Disease Classification on Chest Radiographs}
\runningtitle{Data-Efficient Vision Transformers for Classification on Chest Radiographs}

\author*[1]{Finn Behrendt}
\author[2]{Debayan Bhattacharya}
\author[3]{Julia Krüger} 
\author[3]{Roland Opfer} 
\author[2]{Alexander Schlaefer} 
\runningauthor{Finn Behrendt et al.}

\affil[1]{\protect\raggedright 
  Institute of Medical Technology and Intelligent Systems, Hamburg University of Technology,
 e-mail: finn.behrendt@tuhh.de}
\affil[2]{\protect\raggedright
  Institute of Medical Technology and Intelligent Systems, Hamburg University of Technology}
\affil[3]{\protect\raggedright 
  Jung Diagnostics GmbH, Hamburg, Germany}
\abstract{
Radiographs are a versatile diagnostic tool for the detection and assessment of pathologies, for treatment planning or for navigation and localization purposes in clinical interventions. However, their interpretation and assessment by radiologists can be tedious and error-prone. Thus, a wide variety of deep learning methods have been proposed to support radiologists interpreting radiographs. \\Mostly, these approaches rely on convolutional neural networks (CNN) to extract features from images. Especially for the multi-label classification of pathologies on chest radiographs (Chest X-Rays, CXR), CNNs have proven to be well suited.
On the Contrary, Vision Transformers (ViTs) have not been applied to this task despite their high classification performance on generic images and interpretable local saliency maps which could add value to clinical interventions. ViTs do not rely on convolutions but on patch-based self-attention and in contrast to CNNs, no prior knowledge of local connectivity is present. While this leads to increased capacity, ViTs typically require an excessive amount of training data which represents a hurdle in the medical domain as high costs are associated with collecting large medical data sets.\\ In this work, we systematically compare the classification performance of ViTs and CNNs for different data set sizes and evaluate more data-efficient ViT variants (DeiT). Our results show that while the performance between ViTs and CNNs is on par with a small benefit for ViTs, DeiTs outperform the former if a reasonably large data set is available for training.}

\keywords{Deep Learning, Chest Radiograph, Vision Transformer, Convolutional Neural Network, CheXpert}

\maketitle

\section{Introduction} 
Chest radiographs (CXR) are commonly used for the identification, assessment and localization of pathologies. CXRs enable a cost- and time-effective examination with low radiation dose and allow clinicians to detect a wide range of diseases, plan treatments and localize specific anatomic structures. Therefore, CXRs are the most performed imaging study with an annually increasing number of examinations \cite{united2008effects,raoof2012interpretation,CALLI2021102125,brady_error_2017}. A direct consequence of the increasing amount of CXR examinations is a significantly increased workload for radiologists. Therefore, radiologists need to assess a large amount of CXRs manually in their daily routine which can lead to an increased amount of human-errors \cite{brady_error_2017,Berlin2007}. Thus, a well-integrated computer-assisted tool that could give cues to the radiologists on what pathology might be present and where to look, could accelerate clinical workflows and reduce the number of human errors. Furthermore, such systems could be especially helpful for inexperienced radiologists and help to prioritize assessments of CXRs \cite{CALLI2021102125}.
Various computer-assisted tools, including feature engineering and later statistical models that learn from training data have been proposed in the past for this task. 
Finally, the publication of large-scale data sets such as CheXpert or MIMIC-CXR \cite{10.1609/aaai.v33i01.3301590,johnson_mimic-cxr_2019} paved the way towards human-level classification performance on CXRs with deep-learning based CNNs \cite{10.1609/aaai.v33i01.3301590,CALLI2021102125,LITJENS201760}.
Furthermore, CNNs are proposed for a wide variety of tasks such as classification, localization, segmentation or automated report generation and have emerged to be the de-facto standard for the processing of radiographs. However, a recent publication challenges CNNs and proposes Vision Transformers (ViT) that use multi-headed self-attention between image patches instead of convolutions to learn meaningful feature representations from images \cite{dosovitskiy2021an}. Originally, transformer networks have shown strong performance for modeling and interpreting sequence-data like sentences and outperform traditional recurrent neural networks in many sequence-related tasks \cite{NIPS2017_3f5ee243}. Applying the core principles of Transformers to the image domain as it is done in ViTs has shown to outperform plain CNNs for large-scale databases of generic images such as ImageNet. \\ Beside the potential performance gains, ViTs share the appealing property of class-level local attention maps \cite{NIPS2017_3f5ee243}. These attention maps could be helpful not only for the classification task in CXRs but also for tasks where localization of anatomical structures is required. However, ViTs do not impose prior knowledge of the local connectivity of image pixels as it is the case with convolutions. Thus, ViTs require an excessive amount of training data and are often only applicable when pre-trained on large-scale data sets \cite{pmlr-v139-touvron21a}. \\ This opens the question, whether ViTs can be leveraged for multi-class classification problems with CXRs. As large-scale data sets are crucial for pre-training ViTs, it is of interest if their performance improvement against CNNs can also hold in an image domain different from ImageNet with smaller, medical data sets available for fine-tuning.  
In this work, we leverage ViTs for the classification task of pathologies in CXRs and investigate the use of knowledge distillation for data efficiency. 
\\ In summary, our contribution is three-fold: 
\begin{itemize}
    \item We investigate the use of ViTs for multi-label classification in CXRs and compare their performance to CNNs.
    \item We study if knowledge distillation with data-efficient Vision Transformers (DeiT) \cite{pmlr-v139-touvron21a} can improve the classification performance.
    \item We systematically compare the effect of varying training set sizes for CNNs, ViTs and DeiTs, respectively.  
\end{itemize}

\section{Methods} 

\subsection{Data Set} 
We use the publicly available CheXpert Data set \cite{10.1609/aaai.v33i01.3301590}. The data set consists of 224316 CXRs of 65240 patients together with 14 labels, that are automatically generated from radiology reports. There are three types of auto-generated labels. The labels 0 and 1 indicate positive and negative labels, respectively. The third label -1, denotes an uncertain decision. In this work, we treat all uncertain samples as positive samples. Further, we focus on the classification of five different pathologies, namely \textit{Atelectasis}, \textit{Cardiomegaly}, \textit{Consolidation}, \textit{Edema} and \textit{Pleural Effusion}. We split our data into a train, validation and test set. 20\% of the data are used for evaluation ($\mathcal{D}_{test}$). From the remaining data, we sample 5 data folds, each consisting of 80\% training data ($\mathcal{D}^i_{train}$) and 20\% validation data ($\mathcal{D}^i_{val}$), where $i$ indicates the fold. To simulate varying training set sizes we sample subsets including \{10,20,...,90\}\% of $\mathcal{D}^i_{train}$ for all folds respectively.
\\ We do not perform a specific pre-processing of the images but resize the images to a resolution of $224\times224$px. For data augmentation, we apply random augment \cite{NEURIPS2020_d85b63ef} and random erasing.
\subsection{Deep Learning Models} 
For our experiments, we utilize DenseNets \cite{huang2019convolutional} as baseline CNNs as they have proven to be a strong baseline for the classification task on CXRs \cite{CALLI2021102125}. In general, CNNs utilize blocks of convolutions, together with a normalization, non-linear activation functions and pooling operations stacked on each other to map an input image to a feature vector. A linear layer maps the feature vector to the output vector which is compared with the class labels. DenseNets add specific skip connections between the convolutional blocks to allow training deep stacks of these blocks \cite{huang2019convolutional}. We compare different versions of the baseline CNN, namely DenseNet-121 and DenseNet-201 where the main difference is the depth of the architecture and thus, the number of trainable parameters.  \\
In contrast to CNNs, ViTs do not process image arrays by convolutions. Instead, the image $X \in \mathrm{R}^{H\times W \times C}$ is cropped into $N$ patches $x_p \in \mathrm{R}^{N\times(P^2\times C)}$, where $H,W$ is the dimension of the image, $C$ is the number of channels and $P$ the resolution of the cropped patches. The patches $x_p$ are flattened and mapped to a fixed dimension $D$ by a linear layer. Additionally, a class token is prepended to the mapping which is later used as input for a classification layer. Furthermore, A 1-dimensional position embedding is added to each patch embedding. The resulting sequence of image patches, class tokens and positional embeddings is used as input to the encoder of the ViT. The encoder includes multiple stacked transformer blocks. Each block consists of a multi-headed self-attention and a multilayer perceptron layer with a normalization layer and skip-connections in between. \\ 
We include different versions of ViTs in our experiments, namely ViT-Small (ViT-S) and ViT-Base (ViT-B). The differences between the versions are the number of encoder layers, the dimension of the embeddings $D$, the MLP configuration and the number of attention heads \cite{NIPS2017_3f5ee243}. \\
We further include data-efficient Vision Transformers (DeiT) \cite{pmlr-v139-touvron21a} to our study. DeiTs share the same overall architecture as ViTs. In addition to the class token, a distillation token is added to the patch embedding. Similar to the class token, the distillation token interacts with the patch embeddings through self-attention in the encoder blocks and is processed by a classification layer to obtain an output vector. It is used in a knowledge distillation framework, where the Kullback-Leibler divergence between the output of a teacher network and the output of the distillation token is added to the loss function together with the loss between the class token and the ground truth. The authors of \cite{pmlr-v139-touvron21a} speculate that by this, the inductive bias of CNNs can be distilled to ViTs, which makes DeiTs more data-efficient compared to plain ViTs. \\ We include the pre-trained versions Deit-S and Deit-B in our studies and investigate two use-cases of DeiTs. First, we use pre-trained DeiT networks that apply the knowledge distillation process only during pre-training on ImageNet. Second, we investigate using knowledge distillation with a trained DenseNet-201 as a teacher network during fine-tuning on the CheXpert data set. These distilled models are denoted as Deit-S-Dist and Deit-B-Dist, respectively. \\
We train our networks for a maximum number of 50 epochs and use early stopping based on the validation loss. We use binary cross-entropy loss as a loss function with inverse frequency weighting to account for the class imbalance in the training data. For both, CNNs and ViTs, we use AdamW as optimizer and a batch size of 128. We scale our learning rate with a cosine schedule and use two warmup epochs where we linearly increase the learning rate. While we use an initial learning rate of $lr=0.0001$ for CNNs, ViTs require a smaller initial learning rate of $lr=0.00005$. We search the hyperparameters based on the performance on the validation set $\mathcal{D}_{val}$.

\begin{figure}[t]
\includegraphics[scale=.315]{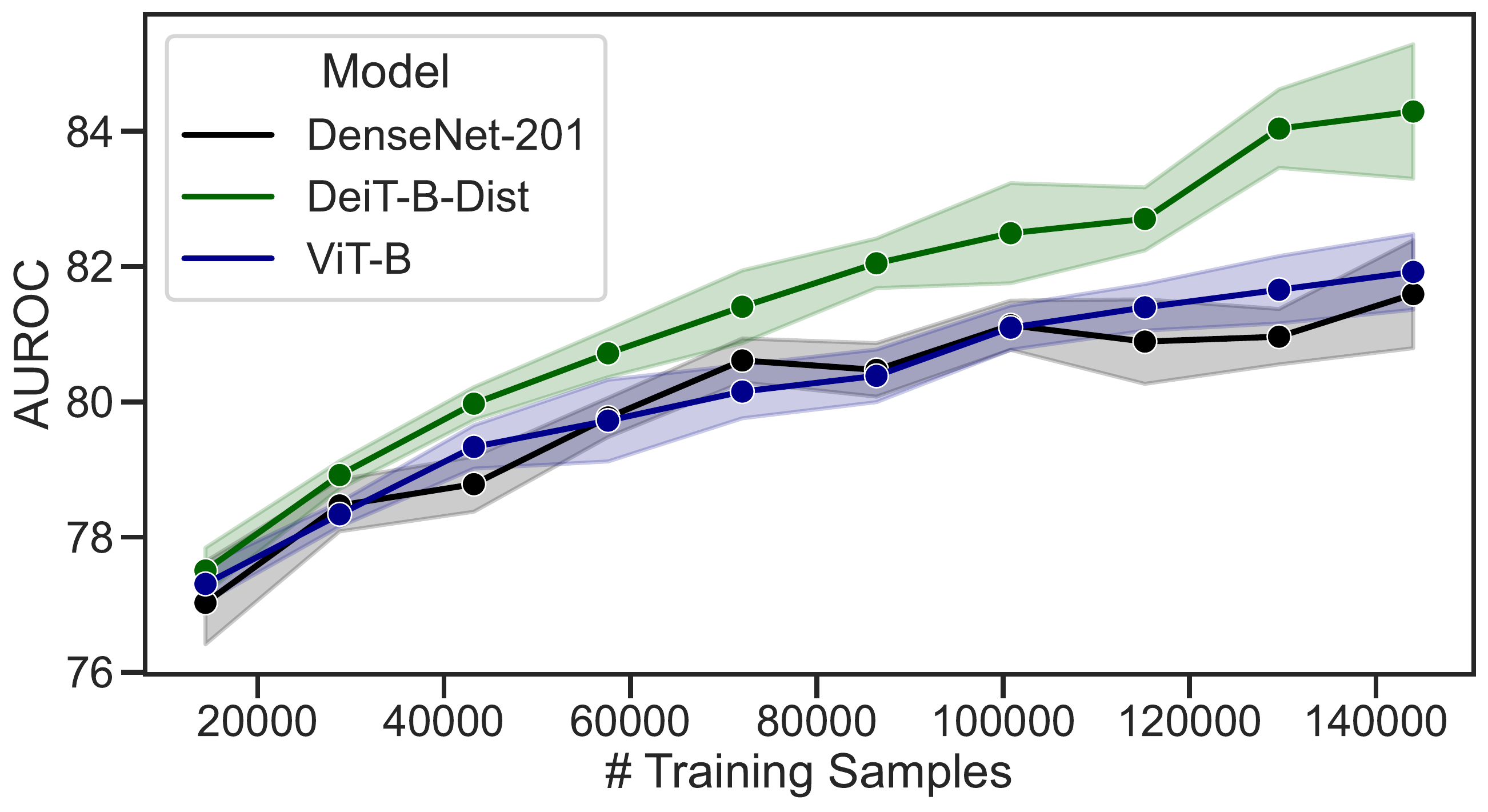}
\caption{Classification performance for different proportions of the CheXpert data set as training set. The average AUROC values of the 5-Fold cross-validation are reported in percent. Standard deviations are visualized as enveloping intervals. }
\label{fig:setsize}
\end{figure}

\section{Results} 
\begin{figure}[t]
\includegraphics[scale=.4]{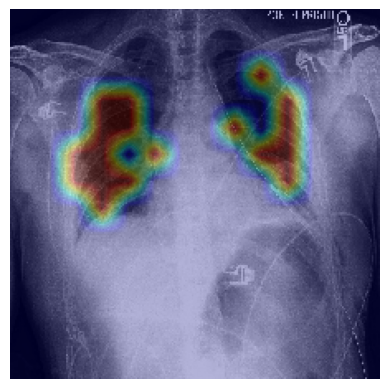}
\includegraphics[scale=.4]{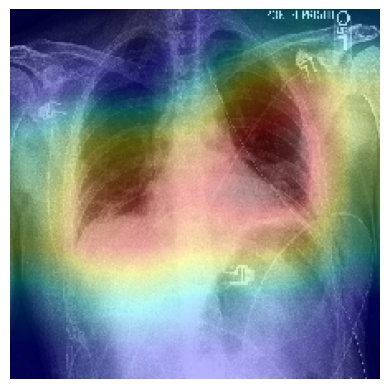}
\caption{Attention visualization for DeiT-B (left) and DenseNet-B (right). The exemplary shown image is labelled with \textit{Atelectasis} and \textit{Pleural Effusion}. For DeiT-B, the attention map is directly accessed form the last layer and interpolated to the image dimension. For DenseNet-B, Grad-Cam is used to generate the attention visualization.}
\label{fig:cam}
\end{figure}

We report the Area under Receiver Operator Curve (AUROC) and the F1 score to evaluate the classification performance. Both metrics are calculated as weighted averages over the five different pathology classes, where each class is weighted by the number of true instances for each label. We report the average performance of the 5-Fold cross-validation together with the standard deviation. \\
As shown in Table \ref{tab:results}, it can be observed that ViT models are on par with the DenseNet baselines. Notably, DenseNet-121 shows competitive performance to ViT-B while requiring significantly fewer parameters. Considering DeiT, both variants show superior classification performance compared to DenseNet and ViT. Comparing Deit-B and Deit-B-Dist, similar classification performance can be observed. \\
Figure \ref{fig:setsize} shows that for all models the data set size has a crucial impact on the classification performance. For DeiT-B-Dist, a higher performance gain can be observed compared to DenseNet-201 and ViT-B especially when training with larger training sets. Overall, it can be observed that even for small data set sizes, the transformer-based models show similar performance compared to DenseNets. \\
To visualize the pixel-wise attention of the networks, saliency maps are provided in Figure \ref{fig:cam}. For DenseNets, a Grad-Cam approach is used to visualize the attention. For transformer-based models, the self-attention weights are visualized. It is noticeable that both networks attend to meaningful regions in the CXR. While the visualization of the attention map weights of transformers leads to local attention maps, Grad-CAM-based saliency maps rather highlight coarse regions.  
\begin{table}[t]
\begin{tabular}{lccc}
Model &     F1 &            AUROC & Param. ($10^6$) \\
\midrule
DenseNet-121 & 63.05$ \pm $0.77 & 81.91$ \pm $0.56 &             6.96 \\
DenseNet-201 & 62.79$ \pm $0.62 & 81.59$ \pm $0.71 &             18.10 \\
       ViT-S & 62.67$ \pm $0.24 & 81.79$ \pm $0.38 &             21.67 \\
       ViT-B & 62.32$ \pm $0.39 &  81.92$ \pm $0.50 &            85.80 \\
      DeiT-S & 63.85$ \pm $0.93 &  83.02$ \pm $0.70 &            21.67 \\
      DeiT-B & 64.93$ \pm $0.88 &  84.02$ \pm $0.90 &            85.81 \\
      DeiT-S-Dist & 63.97$ \pm $1.17 & 82.73$ \pm $1.06 &        21.67 \\
      \textbf{DeiT-B-Dist} & \textbf{65.51$ \pm $0.79} & \textbf{84.56$ \pm $0.91} &        85.81 \\
\end{tabular}
\caption{Classification Performance on the CheXpert Data set. AUROC and F1 scores are provided in percent and the average of 5 cross-validation folds is reported together with the standard deviation. Distilled indicates if a model is trained with knowledge distillation from a DenseNet-201 teacher. Param. denotes the number of trainable parameters in million. The suffix *-Dist denotes models that are fine-tuned with knowledge distillation with DenseNet201 as a teacher.}
\label{tab:results}
\end{table}
\section{Discussion and Conclusion} 
Recently, ViTs show performance gains over classical CNNs on generic images from benchmark data sets such as the ImageNet data set. Furthermore, they add appealing properties like directly accessible and local attention maps. However, due to the missing inductive bias and the exceeding number of trainable parameters, training ViTs requires large-scale data sets \cite{pmlr-v139-touvron21a}.  \\
In this work, we investigate if we can utilize ViT models for multi-label classification on CXR images and compare their performance to a baseline CNN. We investigate the effect of different data set sizes and explore if  knowledge distillation can make the training more data-efficient. 
\\
Our results indicate, that the amount of available training data might not be sufficient to reveal the true power of ViT models. We assume that for ViTs, increasing performance will occur at even larger data sets that are not included in this study.  In contrast to that, the more data-efficient DeiT model shows increasing performance already for smaller training sets. While we can conclude that the distillation process makes the training more data-efficient, it is hard to verify if the data efficiency is achieved by mimicking the inductive bias of the teacher CNN \cite{pmlr-v139-touvron21a}. Furthermore, even though the required amount of labelled training data is reduced, still, large data sets are required to achieve performance improvements over CNNs with transformer-based networks. However, regularizing the training by knowledge distillation shows to be beneficial and can help to efficiently train transformer-based models. \\ 
Besides the improved performance of the transformer-based models, they show local and dense saliency patterns. This observation indicates that the attention maps of transformers can be helpful for the localization of lung diseases from CXRs and have the potential to guide treatment planning. 
\\
Overall, we show that self-attention-based ViT models can be valuable alternatives for multi-label pathology classification, especially in combination with knowledge distillation. \\Our results motivate the research on combinations of CNNs that enforce local connectivity priors and highly expressive ViTs with global attention. This could be a promising direction, especially for the application of ViTs in the medical domain, where annotated data sets are typically small.
\\\\
\textsf{\textbf{Author Statement}}\\
Research funding: This work was partially funded by Grant Number KK5208101KS0. 
\\Conflict of interest: Authors state no conflict of interest.

\bibliographystyle{amsplain}
\bibliography{refs.bib}

\providecommand{\bysame}{\leavevmode\hbox to3em{\hrulefill}\thinspace}
\providecommand{\MR}{\relax\ifhmode\unskip\space\fi MR }
\providecommand{\MRhref}[2]{%
  \href{http://www.ams.org/mathscinet-getitem?mr=#1}{#2}
}
\providecommand{\href}[2]{#2}
\begin{thebibliography}{10}

\bibitem{Berlin2007}
Leonard Berlin, \emph{Accuracy of diagnostic procedures: Has it improved over
  the past five decades?}, AJR. \textbf{188} (2007), 1173--8.

\bibitem{brady_error_2017}
Adrian~P. Brady, \emph{Error and discrepancy in radiology: inevitable or
  avoidable?}, Insights into Imaging \textbf{8} (2017), no.~1, 171--182 (eng).

\bibitem{NEURIPS2020_d85b63ef}
Ekin~Dogus Cubuk, Barret Zoph, Jon Shlens, and Quoc Le, \emph{Randaugment:
  Practical automated data augmentation with a reduced search space}, NIPS
  2020, vol.~33, Curran Associates, Inc., 2020, pp.~18613--18624.

\bibitem{dosovitskiy2021an}
Alexey Dosovitskiy, Lucas Beyer, Alexander Kolesnikov, Dirk Weissenborn,
  Xiaohua Zhai, Thomas Unterthiner, Mostafa Dehghani, Matthias Minderer, Georg
  Heigold, Sylvain Gelly, Jakob Uszkoreit, and Neil Houlsby, \emph{An image is
  worth 16x16 words: Transformers for image recognition at scale}, ICLR, 2021.

\bibitem{huang2019convolutional}
Gao Huang, Zhuang Liu, Geoff Pleiss, Laurens Van Der~Maaten, and Kilian
  Weinberger, \emph{Convolutional networks with dense connectivity}, IEEE PAMI
  (2019), 1--1.

\bibitem{10.1609/aaai.v33i01.3301590}
Jeremy Irvin, Pranav Rajpurkar, Michael Ko, Yifan Yu, Silviana Ciurea-Ilcus,
  Chris Chute, Henrik Marklund, Behzad Haghgoo, Robyn Ball, Katie Shpanskaya,
  Jayne Seekins, David~A. Mong, Safwan~S. Halabi, Jesse~K. Sandberg, Ricky
  Jones, David~B. Larson, Curtis~P. Langlotz, Bhavik~N. Patel, Matthew~P.
  Lungren, and Andrew~Y. Ng, \emph{Chexpert: A large chest radiograph dataset
  with uncertainty labels and expert comparison}, AAAI'19/IAAI'19/EAAI'19, AAAI
  Press, 2019.

\bibitem{johnson_mimic-cxr_2019}
Alistair E.~W. Johnson, Tom~J. Pollard, Seth~J. Berkowitz, Nathaniel~R.
  Greenbaum, Matthew~P. Lungren, Chih-Ying Deng, Roger~G. Mark, and Steven
  Horng, \emph{{MIMIC}-{CXR}, a de-identified publicly available database of
  chest radiographs with free-text reports}, Scientific Data \textbf{6} (2019),
  no.~1, 317 (eng).

\bibitem{LITJENS201760}
Geert Litjens, Thijs Kooi, Babak~Ehteshami Bejnordi, Arnaud Arindra~Adiyoso
  Setio, Francesco Ciompi, Mohsen Ghafoorian, Jeroen~A.W.M. {van der Laak},
  Bram {van Ginneken}, and Clara~I. Sánchez, \emph{A survey on deep learning
  in medical image analysis}, Med. Image Anal. \textbf{42} (2017), 60--88.

\bibitem{united2008effects}
United Nations Scientific~Committee on~the Effects~of Atomic~Radiation et~al.,
  \emph{Effects of ionizing radiation}, Scientific Annexes E (2008), 203--204.

\bibitem{raoof2012interpretation}
Suhail Raoof, David Feigin, Arthur Sung, Sabiha Raoof, Lavanya Irugulpati, and
  Edward~C Rosenow~III, \emph{Interpretation of plain chest roentgenogram},
  Chest \textbf{141} (2012), no.~2, 545--558.

\bibitem{pmlr-v139-touvron21a}
Hugo Touvron, Matthieu Cord, Matthijs Douze, Francisco Massa, Alexandre
  Sablayrolles, and Herve Jegou, \emph{Training data-efficient image
  transformers; distillation through attention}, ICML, vol. 139, July 2021,
  pp.~10347--10357.

\bibitem{NIPS2017_3f5ee243}
Ashish Vaswani, Noam Shazeer, Niki Parmar, Jakob Uszkoreit, Llion Jones,
  Aidan~N Gomez, \L~ukasz Kaiser, and Illia Polosukhin, \emph{Attention is all
  you need}, NIPS 2017, vol.~30, Curran Associates, Inc., 2017.

\bibitem{CALLI2021102125}
Erdi Çallı, Ecem Sogancioglu, Bram {van Ginneken}, Kicky~G. {van Leeuwen},
  and Keelin Murphy, \emph{Deep learning for chest x-ray analysis: A survey},
  Med. Image Anal. \textbf{72} (2021), 102125.

\end{thebibliography}

\end{document}